\pdfoutput=1

\documentclass[11pt]{article}
\usepackage[table]{xcolor}
\usepackage[]{emnlp2021}
\usepackage{multirow}
\usepackage{multicol}
\usepackage{graphics}
\usepackage{graphicx}
\usepackage{caption}
\usepackage{subcaption}
\usepackage{times}
\usepackage{latexsym}
\usepackage{xspace}
\usepackage{hyperref}
\usepackage{booktabs}
\usepackage{verbatim}
\newcommand{\mil}{machine-in-the-loop }
\newcommand{\Mil}{Machine-in-the-loop }
\newcommand{\MiL}{Machine-in-the-Loop\xspace}

\newcommand{\modelname}{Creative Rewriting Assistant\xspace}
\newcommand{\modelacc}{CRA\xspace}
\usepackage{array}
\newcommand{\PreserveBackslash}[1]{\let\temp=\\#1\let\\=\temp}
\newcolumntype{C}[1]{>{\PreserveBackslash\centering}p{#1}}
\newcolumntype{R}[1]{>{\PreserveBackslash\raggedleft}p{#1}}
\newcolumntype{L}[1]{>{\PreserveBackslash\raggedright}p{#1}}

\newcommand{\eg}{e.g.,\xspace}
\newcommand{\ie}{i.e.\xspace}

\newcommand{\vis}[1]{\textcolor{green}{[Vis: #1]}}
\renewcommand\vis[1]{}

\usepackage[T1]{fontenc}

\usepackage[utf8]{inputenc}

\usepackage{microtype}

\usepackage{cleveref}
\usepackage{enumitem}

\title{Machine-in-the-Loop Rewriting for Creative Image Captioning}

\author{Vishakh Padmakumar \\
  New York University \\
  \texttt{vishakh@nyu.edu} \\\And
  He He \\
  New York University \\ 
  \texttt{hehe@cs.nyu.edu} \\}

\begin{document}
\maketitle
\begin{abstract}

Machine-in-the-loop writing aims to build models that assist humans to accomplish their writing tasks more effectively. Prior work has 
found that providing users a machine-written draft or sentence-level continuations has limited success since the generated text tends to deviate from users' intention.  
To allow the user to retain control over the content,
we train a rewriting model that, when prompted, modifies specified spans of text within the user's original draft to introduce descriptive and figurative elements 
in the text.
We evaluate the model on its ability to collaborate with humans on the task of creative image captioning. 
On a user study through Amazon Mechanical Turk,
our model is rated to be more helpful by users than a baseline infilling language model.
In addition, third-party evaluation shows that users write more descriptive and figurative 
captions when collaborating with our model compared to completing the task alone.
However, the improvement is not uniform across user groups:
the model is more helpful to skilled users, which risks widening the gap between skilled and novice users,
highlighting a need for careful, user-centric evaluation of interactive systems.\footnote{Our code and pretrained models are available at \url{https://github.com/vishakhpk/mil-creative-captioning}}

\end{abstract}

\section{Introduction}
\label{sec:intro}

Creative writing tasks are challenging for humans because of their open-ended nature. 
Prior work shows that exposing authors to a collaborator that provides independent suggestions  
can spark new ideas \cite{garfield2008creativity}. This has motivated a line of work in \mil writing \cite{clark2018creative, roemmele2015creative, samuel2016design} where a human collaborates with a model to complete a writing task.
However, recent work 
has shown that providing humans a draft 
generated by a machine is not very helpful
because it may diverge from the direction envisioned by the author \cite{clark2018creative}.
As a result, very little machine-generated text is ultimately retained \cite{akoury2020storium}. 

In this work, we aim to provide a form of interaction that gives human authors more control over the content  
while also assisting them to better express their own ideas \cite{roemmele2015creative}.
We focus on the setting where authors have a clear writing outline but would benefit from suggestions on wording or framing.
To allow authors to control the content,
we develop a \mil system called \modelname (\modelacc) which either rewrites a span of text
or infills between two pieces of text when requested (\Cref{fig:mil_paradigm}). \modelacc is a sequence-to-sequence model, 
building upon recent advances in controllable text generation \cite{shih2019xl, ma2020powertransformer, kumar2020iterative} and
text infilling \cite{donahue2020infilling, fedus2018maskgan, joshi2019spanbert, shen2020blank}.
We train the \modelacc model on a pseudo-parallel corpus of sentence pairs---a generic sentence and a more descriptive or figurative alternative (\Cref{sec:model_train}).

\definecolor{gold}{rgb}{1.0, 0.8, 0.21}
\begin{figure*}[ht]
    \centering
    \includegraphics[width=\textwidth]{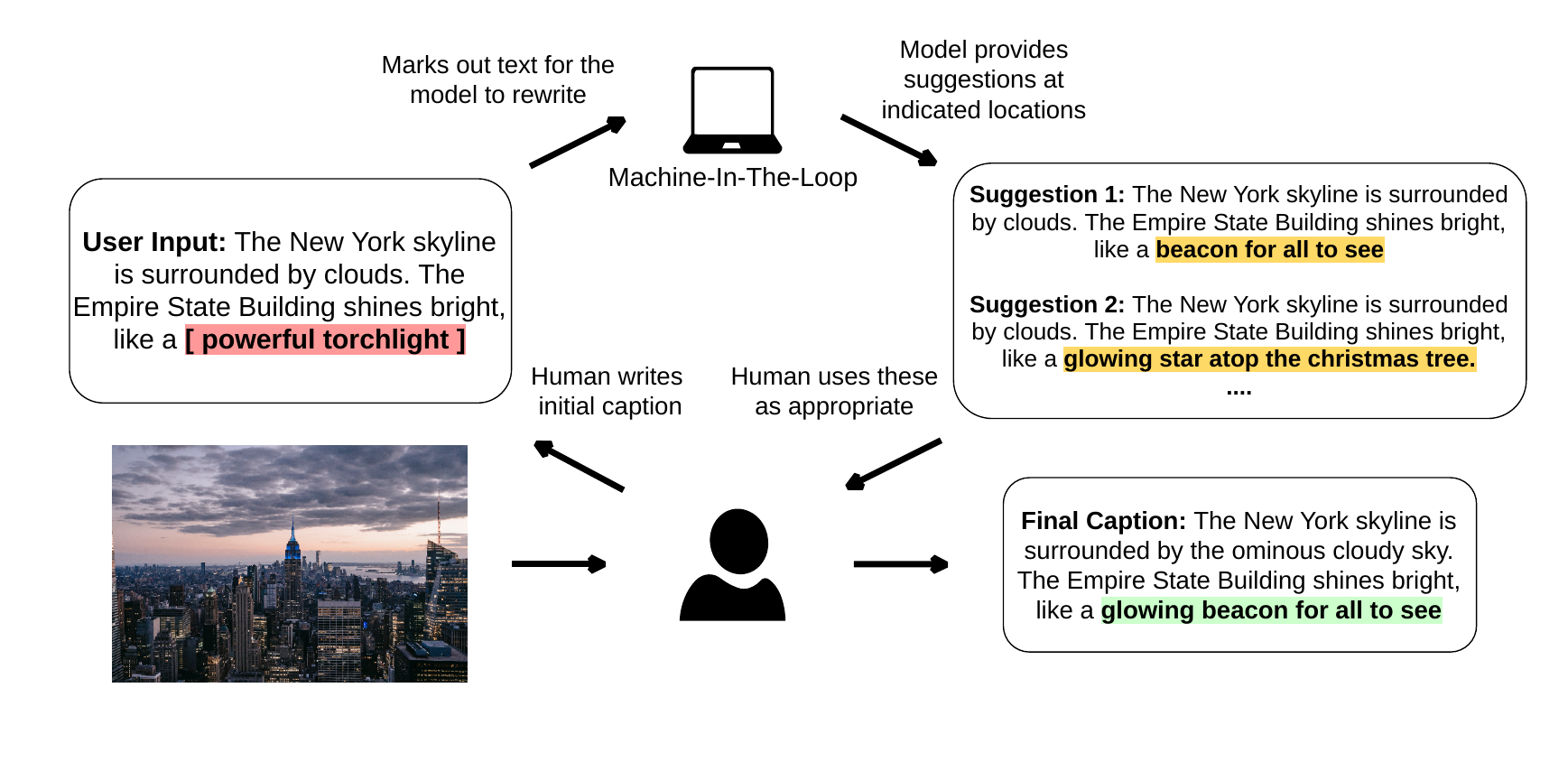}
    \caption{
    \Mil rewriting for image captioning. The human is the central actor in the writing process and initiates interactions with the model by indicating what \colorbox{red!30}{spans of text are to be rewritten}. The model provides \colorbox{gold!50}{suggestions} at these locations and the user chooses \colorbox{green!30}{how to use them}.}
    \label{fig:mil_paradigm}
\end{figure*}

We evaluate the model 
on the task of creative image captioning (\Cref{sec:cic}). 
Users that collaborate with \modelacc report that it is significantly more helpful than a baseline infilling language model (\Cref{sec:user_eval}). Additionally, through a controlled experiment, we find that, on average, users writing with \modelacc produce more creative captions than those writing without assistance, highlighting the end-to-end benefit of our \mil setup (\Cref{sec:third_party_eval}). In particular, users writing with the model produce captions with a more diverse vocabulary. 

To understand how the system impacts different users, 
we analyze the user-model interaction logs (\Cref{sec:analysis}) and find that 
the \mil setup is more helpful to skilled writers because they tend to request targeted suggestions for shorter spans of text while giving the model sufficient context. 
This highlights a need to study the impact of 
study of interactive systems on different user groups as these become more ubiquitous in assisting content generation  given that this technology could result in an even wider gap in performance between different users.

\section{System Overview}
\label{sec:mil}
\begin{figure*}
    \centering
    \includegraphics[width=\textwidth]{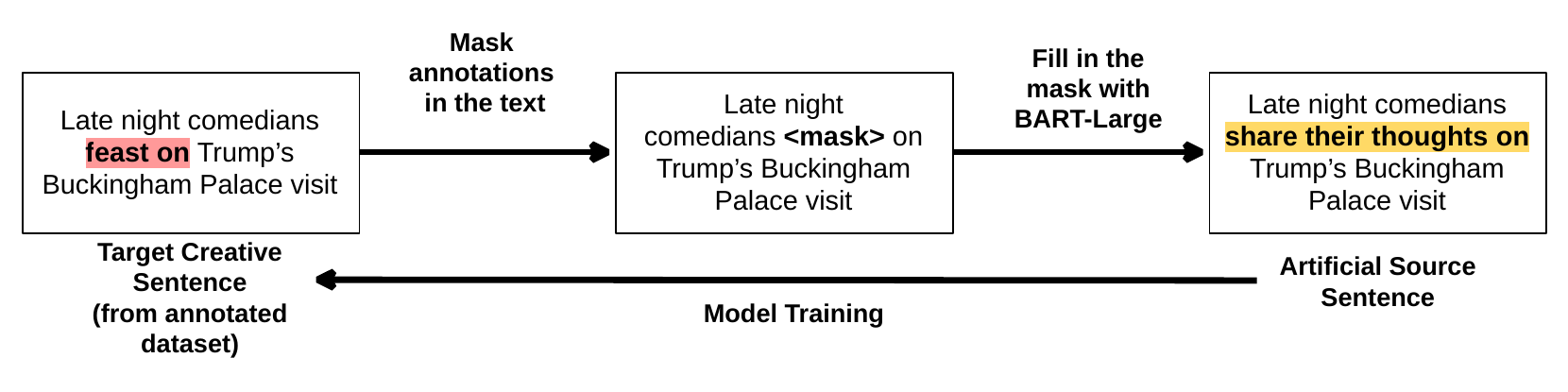}
    \caption{Training data creation. The source sentence is created by masking out the \colorbox{red!30}{annotated span} and \colorbox{gold!50}{infilling it using BART-Large}. The model is then trained to produce the creative sentence from the synthesized source sentence.}
    \label{fig:model_training}
\end{figure*}

\begin{table*}[ht!]
    \centering
    \footnotesize
    \begin{tabular}{|p{3cm}|p{2.5cm}|p{2.5cm}|p{6cm}|}
      \hline
        \textbf{Source} & \textbf{Domain} & \textbf{Annotation} & \textbf{Example} \\
      \hline
        \citet{mohammad2016metaphor} & WordNet example sentences & Words that elicit emotion & I \textbf{attacked} the problem as soon as I was up.\\
      \hline    
        \citet{gordon2015corpus} & Text collected by \citet{mohler2015cross} &
        Metaphors in text
        & I will be out in the city today, feeling the vinous veinous thrust of blood, \textbf{the apple-red circulation of democracy}, its carnal knowledge without wisdom.\\
      \hline
        \citet{bostan2020goodnewseveryone} & Headlines &  Textual cues associated with emotion  & Detention centers will \textbf{shock the conscience} of the nation.\\
      \hline
        \citet{niculae2014brighter} & 
        Product reviews &
        Figurative language & The stones appeared dull and almost opaque, \textbf{like black onyx}, with none of the sparkle you would expect from something called a diamond. \\
      \hline
        \citet{vuamc} & News, fiction and academic text &  Metaphors and personification & \textbf{Like a buzzard in the eyrie, } he would fly around.\\
      \hline
    \end{tabular}
    
    \caption{Sources of creative text and annotations used for creating training examples.}
    \label{tab:sources}
\end{table*}

\paragraph{Creative Image Captioning} 
\label{sec:cic}
To evaluate our system, ideally, we would use tasks like poem or story writing. However, it is challenging to
control the content for a fair comparison of different systems on such open-ended tasks.
Therefore, we evaluate on a proxy task, creative image captioning \cite{chen2015deja}, where the user is asked to write an expressive caption (a figurative or descriptive one as opposed to a literal one) for a given image.
The user is given access to a model that provides editing suggestions while they are working on the task. Our goal is to study if collaborating with the model makes them more effective at completing the task.
Note that our model does not use the image when generating the suggestions,
which is analogous to real use cases where the model does not have access to the author's global writing plan
but instead provide improvements based on the \textit{local} textual context.

\paragraph{\MiL Rewriting} 

An overview of our system is illustrated in \Cref{fig:mil_paradigm}. The user collaborates with the model to complete the writing task. 
We follow the user-initiative setup \cite{clark2018creative}
where the model provides suggestions only when requested by the user.
The system facilitates two types of editing: span rewriting and text infilling.
Given a piece of text (written by the user),
to request span rewriting, the user demarcates spans within the text that need to be rewritten.
The model then edits the marked spans.
For example, given  \textit{``The iPhone was a \textbf{[great piece of technology]} that changed the world''}, the model suggests the rewrite 
\textit{``The iPhone was a \textbf{revolution in technology} that changed the world''}. 
To request text infilling, the user marks blanks to be  infilled.
For example, given \textit{``The lion stalks the deer, a \textbf{\rule{1cm}{0.15mm}}  
in its element''}, the model infills  
\textit{``The lion stalks the deer, a \textbf{predator} in its element''}. 

By limiting the edits to local spans,
we alleviate the issue of deviating from the input content 
or generating incoherent text \cite{holtzman2019curious, wang2020exposure}. 
For both rewriting and infilling, multiple suggestions are provided for the user to consider.
Then, they have the option to either accept a suggestion and continue writing,
or reject them and retain their initial draft.
This interactive process continues until the user is satisfied with the text and indicates the completion of the writing task.

\section{Approach}
\label{sec:approach}

\subsection{Learning from Creative Text}
\label{sec:model_train}

Our goal is to train a model capable of rewriting specific spans of an input sentence requested by a user, to assist them with the creative writing task. 
To this end, we need a dataset of paired sentences where the target sentence is produced by replacing or inserting text spans in the source sentence to make it more descriptive or figurative.
While no such datasets exist,
there are many resources that study creative text by annotating text spans
with their corresponding 
literary devices (including metaphors, emotional cues, and figurative comparisons). 
We thus take the creative text from these datasets as the target sentences,
and synthesize the source sentences by replacing the annotated creative spans with infills from a \emph{generic} language model,
which typically produces less creative text.
An example is shown in Figure \ref{fig:model_training}. %

Specifically, we start with a creative sentence from one of the datasets listed in \Cref{tab:sources},
mask the annotated creative spans in it, and infill them using the pre-trained BART model \cite{lewis2019bart} to generate the non-creative source sentence. 
For each pair from this pseudo-parallel corpus, we create one rewriting example
by inserting the rewrite markers, \texttt{<replace>} and \texttt{</replace>},
at the beginning and the end of the rewritten span, 
as well as one infilling example by replacing the span with a mask token,
\texttt{<mask>}.
We then train \modelacc 
to predict the creative sentences given the generic source sentence using cross-entropy loss. %

\subsection{Learning from Interactions}
\label{sec:learning_from_interactions}
One important advantage of \mil systems is that they can be improved given user feedback.
Once users interact with \modelacc, we obtain their reaction to the suggestions, \ie acceptance and rejection.
This feedback allows us to update the model,
so that it adapts to the observed user preference over time.
Specifically, we create an example pair whenever the user indicates a preference for one sentence over another when presented with model suggestions. 
When the user accepts a suggestion, we take the accepted suggestion as the target (creative) sentence and the user's initial input as the source (non-creative) sentence.
Similarly, when the user rejects a suggestion, we take the rejected suggestion as the source and the user's initial input as the target.
Thus, the model always learns to improve the source sentence.
We then add these new pairs to a similar-sized subset of the original training examples (to prevent forgetting) and fine-tune the rewriting model on the combined dataset. 

\section{Experimental Design}
\label{sec:rq}
We train the \modelacc model using the scheme laid out in \Cref{sec:model_train} and deploy it in the \mil setup detailed in \Cref{sec:mil} in order to answer the following research questions.
\paragraph{User Experience} When we study collaborative writing, the key stakeholder is the user so we evaluate if users writing in tandem with \modelacc find the model helpful. To answer this, we run a user study and compare suggestions obtained from \modelacc and a baseline BART model in the \mil setup (\Cref{sec:user_eval}). Once \modelacc is deployed with real users, we would like to adapt it to users' preferences inferred from the observed interactions. Hence we also compare a user-adapted model (per \Cref{sec:learning_from_interactions}) to the previously deployed \modelacc (\Cref{sec:model_update_eval}). 
\paragraph{Quality of the Writing} We also want to study the outcome of the collaboration. In particular, does \modelacc help users' write higher quality captions for images? We compare captions collected from the \mil setup to those obtained from solo-writers using third-party annotators to see if users write more creative captions in a collaborative setup (\Cref{sec:third_party_eval}). 
\paragraph{Broader Impact of Collaboration} In the \mil setup, we introduce \modelacc into the writing process. This intervention potentially impacts different users differently. 
In particular, we study how skilled and novice user groups interact 
with \modelacc to understand how such a model impacts the skill gap between users of different backgrounds (\Cref{sec:analysis}).  

\section{Experiments}
\label{sec:experiments}

\paragraph{User Study}

We hire users on Amazon Mechanical Turk to perform the creative image captioning task.
A screenshot of our user interface and the details about worker remuneration are provided in \Cref{sec:hit_instructions}. The plan for our user study was approved by the Institutional Review Board of our university. 
Each user is presented with an image and asked to write a caption
that is as figurative and/or descriptive as possible with at least 100 characters. 
The images were randomly sampled from the figurative subset of the D\'ej\`a Captions dataset \cite{chen2015deja}, where the gold caption contains literary elements like metaphors and hyperbole.
We ask users to request suggestions from the model at least twice while they are writing;
however, they are free to ignore the suggestions.
Users are instructed to use square brackets (as seen in \Cref{fig:mil_paradigm}) to mark spans to be rewritten and
underscore to indicate blanks to be infilled.
They can edit the text with the model iteratively until they are satisfied with the caption. 
Once users submit the final caption, they are asked to complete a survey to rate the helpfulness and grammaticality of the assistant
as well as their satisfaction with the final caption.
The full task instruction 
is provided in \Cref{sec:hit_instructions}.

\paragraph{Model Details}

To train the \modelacc 
model, we first create the pseudo-parallel corpus as detailed in \Cref{sec:model_train}. Using creative sentences from all the sources from \Cref{tab:sources}, we obtain a corpus containing 42,000  training pairs, 2,000 validation pairs, and 1,626 test pairs. 
The \modelacc model is trained by fine-tuning the \texttt{fairseq} \cite{ott2019fairseq} 
implementation of BART on the training set of this corpus  
We train the BART-Large pre-trained checkpoint from \texttt{fairseq} 
for 5 epochs
with a learning rate of $3 \times 10^{-5}$. The learning rate was selected by held-out perplexity on the validation set.  We use the recommended default values in \texttt{fairseq} for the hyperparameters  
of the Adam 
optimizer \cite{kingma2014adam}, dropout rate, and learning rate scheduler.\footnote{The beta values for the Adam optimizer are 0.9 and 0.999, the dropout rate is set to 0.1, and we use a polynomial decay learning rate scheduler with the weight decay set to 0.01.} 

To evaluate whether \modelacc provides helpful suggestions, we compare its performance to a pre-trained infilling language model, BART \cite{lewis2019bart}. 
When using BART for rewriting, we mask and then infill the spans of text demarcated by users (regardless of whether they are meant to be rewritten or infilled). 
To produce creative generations, a balance between diversity and fluency is desired during decoding. A small internal pilot shows a lack of diversity in beam search outputs. 
Thus, we use top-$k$ sampling for both models, with $k$ set to 10.


\subsection{User Evaluation of Suggestion Quality}
\label{sec:user_eval}

To evaluate the quality of the suggestions provided by \modelacc vs.\ the pre-trained BART baseline,
we conduct A/B testing on 50 images randomly sampled from the D\'ej\`a Captions dataset. 
Upon connecting to our server, each user is randomly assigned to work with either BART or \modelacc. 
We ensure that each image has one caption from each model. 
In addition, users working with both models are recruited from the same pool during the same time period,
which minimizes the difference in performance due to individual users.

Once the task is completed, we ask the user to answer the following questions about the model on a Likert scale of 1 (worst) to 5 (best): 
\begin{itemize}[itemsep=-1ex,topsep=0.5ex]
    \item How helpful were the model suggestions? 
    \item How grammatically correct were the model suggestions? 
    \item How satisfied were you with the final caption?
\end{itemize}
In addition, to analyze the effect of users' initial writing ability,
we ask them to assess their writing skills:
\begin{itemize}[itemsep=-1ex,topsep=0.5ex]
    \item How would you rate your own writing ability on a scale of 1 to 5? 1---I don't have much experience with writing or am not too confident with the language, to 5---I have writing experience and/or have considerable proficiency with the English language.
\end{itemize}

We also examine if this user-rated helpfulness tallies with automatic metrics that we compute on the observed interactions. 
\paragraph{Results}
The results from the survey are presented in Table \ref{tab:eval_model_vs_bart}. Each reported value is an average of 50 user responses.  
We find that, on average, users find \modelacc to be more helpful than BART and report no significant difference between the two models in terms of grammaticality.  
By training \modelacc on the pseudo-parallel creative corpus, we align the model suggestions better to the creative writing task, resulting in a more helpful collaborator.

To see if the human evaluation tallies with automatic metrics, we calculate the fraction of model suggestions accepted by the users, across the 50 user responses, is reported in \Cref{tab:suggestion_profile}.
\modelacc has a higher acceptance rate than BART,
consistent with the helpfulness rating from users. 
The total number of suggestions requested from BART is slightly higher, perhaps explained by its lower acceptance rate---users may persist with variants upon receiving unsatisfactory suggestions. 

Accepting a suggestion does not necessarily mean that it is useful since the user may
further edit it. In fact, prior work has shown that a large fraction of the suggested text is not retained by the user \cite{akoury2020storium}.
To measure how much of the \emph{accepted} suggestions are actually \textit{used},
we calculated the Rouge-L recall score of accepted suggestions against the final caption submitted by the user.
As shown in \Cref{tab:suggestion_profile}, larger fractions of \modelacc's suggestions were retained by users.

\begin{table}[ht!]
    \centering
    \small
    {
    \begin{tabular}{|l|r|r|}
        \hline 
        \textbf{Question} & \textbf{BART} & \textbf{\modelacc} \\
        \hline
        Helpfulness & 2.23* & \textbf{3.06*} \\
        \hline
        Grammaticality & 2.96 & \textbf{3.22} \\
        \hline
        Satisfaction & \textbf{3.69} & 3.65 \\
        \hline
    \end{tabular}
    }
    \caption{User evaluation of model performance for  pre-trained BART baseline vs. \modelacc. Each value is an average across 50 user scores per model. Bold values  correspond to the higher average. Rows marked with an asterisk indicates statistically significant differences ($p$-value  
    $< 0.05$ according to a Mann-Whitney-Wilcoxon test). 
    Users find the \modelacc model to be more helpful by a statistically significant margin.}
    \label{tab:eval_model_vs_bart}
\end{table}

\begin{table}[h]
    \centering
    \footnotesize
    \begin{tabular}{|L{0.9cm}|R{1.1cm}|R{1.4cm}|R{1.4cm}|R{1.15cm}|}
        \hline 
        & \# request & \# accepted & \% accepted & Rouge-L \\
        \hline
        \textbf{BART} & \textbf{151} & 37 & 24.5 & 0.744 \\
        \hline
        \textbf{\modelacc} & 141 & \textbf{45} & \textbf{31.9} & \textbf{0.824} \\
        \hline

    \end{tabular}
    
    \caption{
    Interaction statistics - How many suggestions were requested and accepted for the different models aggregated across 50 users for each model and the Rouge-L recall scores of accepted model generations against the final caption submitted by the user. The higher score of the two is bolded. 
    Users accept more suggestions and retain more text from CRA.
    }
    \label{tab:suggestion_profile}
\end{table}

\subsection{Effect of Learning from User Interaction}
\label{sec:model_update_eval}

From \Cref{sec:user_eval}, we see that users find \modelacc to be more helpful than an infilling baseline model. In order to further align \modelacc to users' preferences, 
we 
fine-tune the model on paired examples created from their acceptance and rejection of the model suggestions (\Cref{sec:learning_from_interactions}). The interactions with 50 users (collaborating with the \modelacc) 
result in a dataset of 474 pairs of sentences. To ensure that the model does not overfit to these examples and forget prior training on the pseudo-parallel creative corpus 
(\Cref{sec:model_train}), we also sampled  
450
sentence pairs from it and added these to the interaction dataset. 
We then fine-tune the previously deployed \modelacc model for 5 epochs on this dataset. We choose the learning rate of $3 \times 10^{-6}$ using five-fold cross-validation with the criteria of label smoothed cross-entropy loss
\footnote{We again use the released \texttt{fairseq} fine-tuning script retaining the recommended hyperparameters for the Adam optimizer, dropout rate and learning rate scheduler.}.   
We evaluate this \emph{user-adapted} \modelacc model against the initial \modelacc model  
on a fresh sample of 50 images, following the A/B testing setup from \Cref{sec:user_eval}
. 

\paragraph{Does user feedback improve the model?}
Our hypothesis is that adapting the model to user feedback should make it more helpful for subsequent users.  
From Table \ref{tab:eval_init_vs_updated}, we see that 
the users do find the updated model to be slightly more helpful than the initial model on average; however, a Mann-Whitney-Wilcoxon test shows that this difference is not statistically significant ($p$-value $=0.402$). A possible reason is that the different usage patterns of different users leads to the model getting noisy feedback and 
not significantly improving on the initial trained state. 
Thus, 
a potential future direction is to explore adapting the model separately to each user in a few-shot setting, possibly with longer interactions.

\begin{table}[ht!]
    \centering
    \footnotesize
    \begin{tabular}{|L{2.25cm}|R{1.25cm}|R{2.5cm}|}
        \hline 
        \textbf{Question} & \textbf{Initial \modelacc} & \textbf{User-adapted \modelacc} \\
        \hline
        Helpfulness  & 2.81 & \textbf{3.05} \\
        \hline
        Grammaticality  & 2.87 & \textbf{3.26} \\
        \hline
        Satisfaction  & 3.67 & \textbf{3.78} \\
        \hline
    \end{tabular}
    \caption{User evaluation of model performance for the initial model vs. the adapted model trained on user interactions averaged across 50 user scores. Bold values indicate the higher average. Users find the adapted model to be more helpful, although the difference is not statistically significant. 
    }
    \label{tab:eval_init_vs_updated}
\end{table}

\subsection{End-to-End System Evaluation}
\label{sec:third_party_eval}

In \Cref{sec:user_eval}, we observe  
that users find \modelacc 
more helpful than a baseline BART model.
However, is the quality of the caption improved by collaborating with the model? 
To answer this question, we collect three captions for each of the 100 images using three systems: 
two \mil systems using \modelacc and BART respectively, and one solo writing setup. 
For solo writing, we recruit workers from the same pool as before (Amazon Mechanical Turk) and provide them the same instructions as in the \mil setup, except that all mentions of model assistance are removed.
We then ask a `third-party' human annotator (who did not participate in the writing task)
to compare the captions pairwise 
for each image. 
The annotator is presented with two captions for the same image and asked to pick the more creative caption of the two. 
In this manner, we collect 3 separate annotations for each pairwise comparison for each image and decide the winning caption based on a majority vote.  

\paragraph{Does working with \modelacc improve the final caption?}

From \Cref{tab:third_party}, we observe that both collaborative setups (Human+\modelacc and Human+BART) outperformed the solo-writing setup according to the majority vote. %
While prior work in the 
creative domain was unable to match the performance of the human-only baseline using a less controllable %
assistant that provides full length drafts \cite{clark2018creative},
here we show that collaborative setups are able to improve creative output of human users, in-line with the expectations of literature on creativity support systems \cite{garfield2008creativity}. 

\begin{table}[ht!]
\footnotesize
\begin{tabular}{|R{1.9cm}|C{1cm}|C{1cm}|R{1.9cm}|}
\hline
& \multicolumn{2}{c|}{\textbf{Majority Vote Wins}} & \\ \hline
{Human-Only} & 45                 & \textbf{55}                & {Human+BART}  \\
{Human-Only} & 43                 & \textbf{57}                & {Human+CRA}   \\
{Human+BART} & 48                 & \textbf{52}                & {Human+CRA} \\ 
\hline
\end{tabular}
\caption{Pairwise comparison of 100 captions from \mil writing with our model (Human+\modelacc) and the baseline (Human + BART) as well as a human writing without assistance (Human Only). Wins were decided by a majority vote amongst 3 crowd workers. Users write better captions in a collaborative setup.}
\label{tab:third_party}
\end{table}

\begin{figure}[ht!]
\centering
\includegraphics[width=0.35\textwidth]{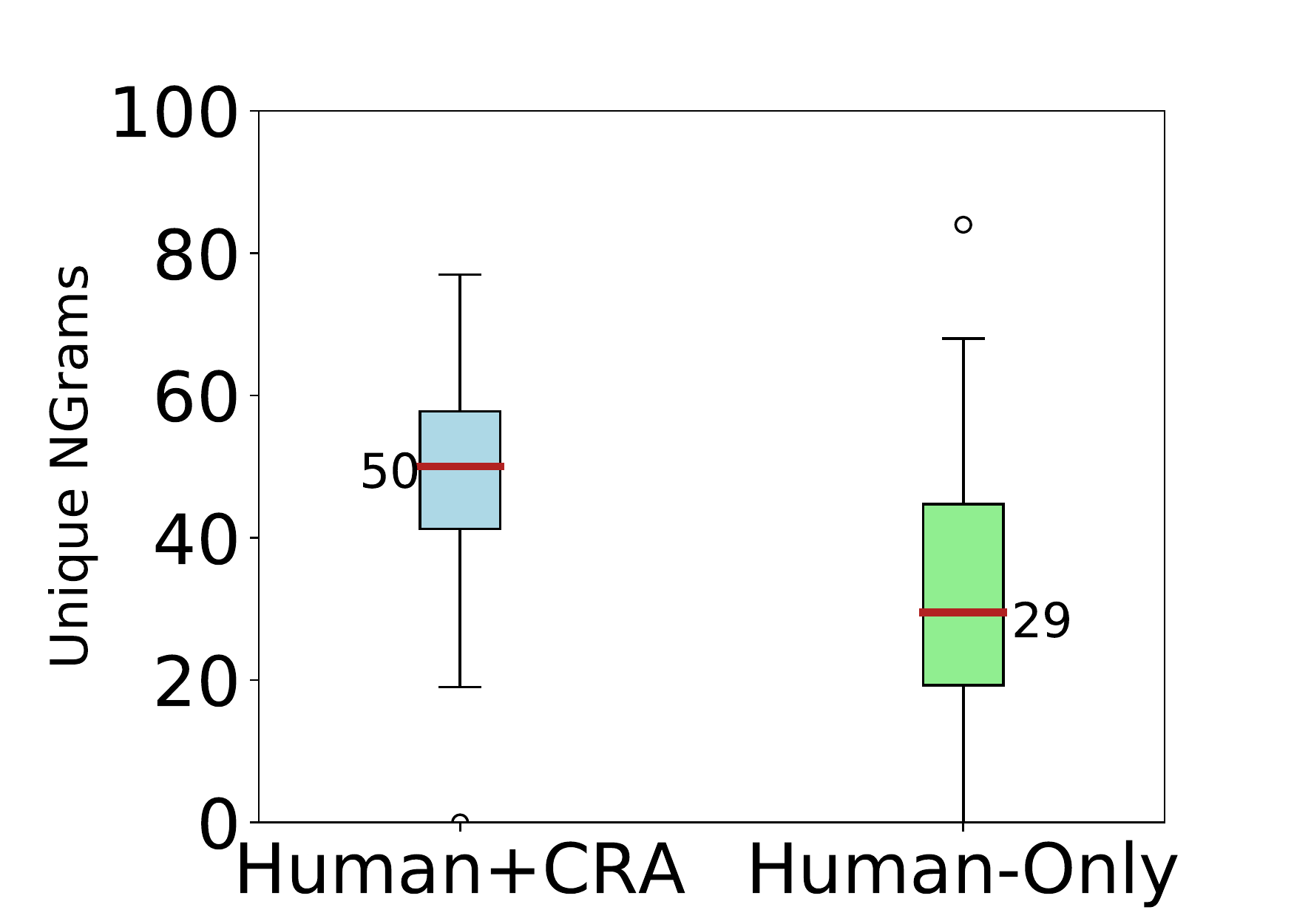}
\caption{Comparison of text generated from a collaborative setup (Human+\modelacc) and solo-writers (Human-Only). Collaborative users tend to write more diverse captions containing more unique trigrams (N=3)
}
\label{fig:cra_human_ngram}
\end{figure}

\paragraph{How does CRA influence the captions? 
}
\label{sec:collabVhuman}
To analyze how model intervention affects the output text, we 
measure the count of unique trigrams 
in 100 captions produced from the Human+\modelacc setup and the Human-Only setup. Collaborative users are exposed to suggestions from an external model so we expect the generated text to contain more diverse vocabulary usage. From \Cref{fig:cra_human_ngram}, we see that, on average, captions generated from the collaborative setup do contain more unique trigrams. 

The improvement in written captions because of the collaboration does not only come from direct model interventions the text. Some users also reported\footnote{We include representative user feedback in \Cref{sec:user_feedback}.} that considering different alternatives suggested by the model provided inspiration on how to improve the text (even though the suggestions are not accepted).

\subsection{Error Analysis}
\label{sec:error_modes}
\begin{table*}[ht!]
    \centering
    \small
    \begin{tabular}{|p{0.2cm}|p{5.8cm}|p{6.3cm}|p{1.8cm}|}
        \hline
         ID & \textbf{Demarcated Source Sentence} & \textbf{Accepted Suggestion} & \textbf{Edit}\\
        \hline
        1 & A solemn woman place her mother's diary on a stepping stone her late father laid in the garden. The \textbf{ [ surrounding pale grass gently sway in the cold breeze ]} while the woman ponders times of the past. Reminiscence now taking over and winter's beginning, the woman braces herself for dreary time to come. 
        & 
        A solemn woman place her mother's diary on a stepping stone her late father laid in the garden. The \textbf{pale grass gently danced and teased in the wind} while the woman pondered times of the past. Reminiscence now taking over and winter's beginning peaks, the woman braces herself for dreary time to come.
        & 
        Figurative language \\
        \hline
        2 & A man walks along the seashore with the horizon looming in the background. The dark clouds \textbf{\rule{0.5cm}{0.15mm}} as the sun sets for the day.
        &   
        A man walks along the seashore with the horizon looming in the background. The dark clouds \textbf{slowly disperse} as the sun sets for the day. 
        &
        Precise wording \\
        \hline
        3 & The image represents wisdom and profound intelligence. It is the face of a man who lead the nation with grace and honor. 
        It is a statue that reflects the \textbf{[ moral aspect of American people today ]}.
        & The image represents wisdom and profound intelligence. It is the face of a man who lead the nation with grace and honor.  
        It is a statue that reflects \textbf{the moral aspect of a great man who lived and breathed the ideals of freedom and democracy}
        &
        Embellishment\\
        \hline
    \end{tabular}
    \caption{Examples  where the model was successful in providing assistance. Bold spans in the source and target sentences are what marked by users and rewritten by the model, respectively.} 
    \label{tab:pos_examples}
\end{table*}
\begin{table*}[ht!]
    \centering
    \footnotesize
    \begin{tabular}{|p{0.2cm}|p{5.9cm}|p{6.5cm}|p{1.4cm}|}
        \hline
         ID & \textbf{Demarcated Source Sentence} & \textbf{Poor Suggestion} & \textbf{Error Type} \\
         \hline 
         1 & In front of a wall, a girl with blonde hair is on her hands who seems to be \textbf{[coming out of a  magical door ]} 
         & In front of a wall, a girl with blonde hair is on her hands who seems to be \textbf{laughing out loud}.
         & 
         Content drift \\
        \hline
        2 & A child stands tall in a \textbf{[ wave ]} on the beach. 
        & A child stands tall in a \textbf{motorized scooter} on the beach.
        & 
        Content drift \\ 
        \hline
        3 & I am witnessing a field of golden grain with a tall flower is blooming.  
        That flower is not yet fully grown, yet its shades of purple are there and plainly visible. \textbf{[ Overall, the image is nice. I do believe, however, that the quality of the image could be sharpened a bit. ]}
        & I am witnessing a field of golden grain with a tall flower is blooming.  
        That flower is not yet fully grown, yet its shades of purple are there and plainly visible. \textbf{Overall, the image is nice. I do believe, however, that the quality of the image could be sharpened a bit.}
        & 
        Repeated the source \\ 
        \hline
        4 & A beautiful \textbf{[ sunset.A ]} beautiful sunset in the ocean lighting up the sky in exotic colors. 
        & A beautiful \textbf{sunset in the ocean lighting up the sky in exotic colors. A breathtaking view of nature at its best.} 
        & 
        Excessive editing \\ 
        \hline
    \end{tabular}
    \caption{
    Examples of rejected model suggestions.  
    Bold spans in the source and target sentences are marked by users and rewritten by the model, respectively. } 
    \label{tab:neg_examples}
\end{table*}

To provide the full picture of \modelacc, we manually labeled 50 rejected suggestions to identify common error modes. Some illustrative examples of these are listed in 
\Cref{tab:neg_examples}. The most common  failure case (21 out of the 50) is content drift: when the model is asked to replace key content words, 
sometimes the rewritten text changes the meaning of the user draft. This is seen in Examples 1 and 2 in \Cref{tab:neg_examples}, where the model changes ``wave'' to ``motorized scooter''; 
while the suggestion is coherent, it changes the original meaning of the sentence. 
This is likely an artifact of how we create the pseudo-parallel corpus of training data: When BART performs infilling, the text introduced is not guaranteed to preserve the original content.\footnote{To validate the quality of the pseudo-paralllel corpus, we randomly sampled 50 sentence pairs and manually checked them for hallucinations. We observed hallucination 4 times out of 50 possibly explaining the observed content drift errors.}
The second common error type (14 out of the 50) is to copy the source text verbatim
(example 3 in \Cref{tab:neg_examples}),
especially when a long text span (\eg a full sentence) is rewritten, 
which is rare in our training data.
Lastly, there is a small fraction of cases (9 out of the 50) when the model makes suggestions outside the desired demarcated region---this is often seen when the demarcated text spans two sentences and contains incoherent phrases (example 4 in \Cref{tab:neg_examples}).

\section{How Does CRA Impact Different 
Users? 
}
\label{sec:analysis} 

\paragraph{Which users find \modelacc more helpful?}
Our main hypothesis is that \modelacc benefits human authors by giving them more control over the global content and providing local wording suggestions \cite{roemmele2015creative}. Thus, its effectiveness relies on the assumption that the user has a coherent writing plan,
which may or may not be true depending on the skill level of the writer.
To analyze the influence of users' inherent writing skill on model effectiveness, we put users into two groups based on their self-assessed writing ability (1 is the least skilled and 5 is the most skilled).
A user is considered a \emph{skilled writer}
if they rate themselves higher than 3
and otherwise a \emph{novice writer}. Out of the 50 users who interacted with \modelacc, 22 fall into the novice group and 28 fall into the skilled group.  As a sanity check, the self-reported skill level is consistent with the result from the third-party evaluation---more captions written by skilled writers were judged as the winning caption than the novice writers (72.72\%  vs. 46.42\%).

We show the ratings of helpfulness of \modelacc and the acceptance rate of model suggestions by user group in \Cref{tab:sa_breakdown}.
We observe that skilled writers find the model more helpful 
and accept a higher fraction of the provided suggestions,
while novice writers tend to request more suggestions with a lower acceptance rate. This is consistent with the hypothesis  
that the skilled writers have a more clear plan thereby playing to the model's strengths. 

To understand if the discrepancy in reported model helpfulness between the two groups is due to them requesting different kinds of suggestions, we identify the characteristics of edits that \modelacc is good at and compare them to the requests made by the two groups.    

\begin{table}[ht!]
    \centering
    \small
    \begin{tabular}{|c|r|r|}
        \hline
        & \textbf{Novice} & \textbf{Skilled} \\ 
         \hline
        Helpfulness  & 2.27* & \textbf{3.23*} \\ 
        \hline
        \# request & \textbf{3.04} & 2.64 \\ 
        \hline
        \% accepted & 29.8 & \textbf{33.7}\\
        
        \hline
    \end{tabular}
    \caption{Breakdown of model performance grouped by self-assessed writing skill. The rows correspond to average ratings of model helpfulness from the user survey, the average number of requests made to the model and the acceptance rate of received suggestions for both user groups. Rows marked with an asterisk indicates statistically significant differences ($p$-value $<  0.05$ on a Mann-Whitney-Wilcoxon test). Bold values correspond to the higher score. Skilled writers find the model significantly more helpful, request fewer suggestions but accept a higher percentage of them.} 
    \label{tab:sa_breakdown}
\end{table}

\begin{figure}[ht!]
\centering
\begin{subfigure}[b]{0.24\textwidth}
    \centering
    \includegraphics[width=\textwidth]{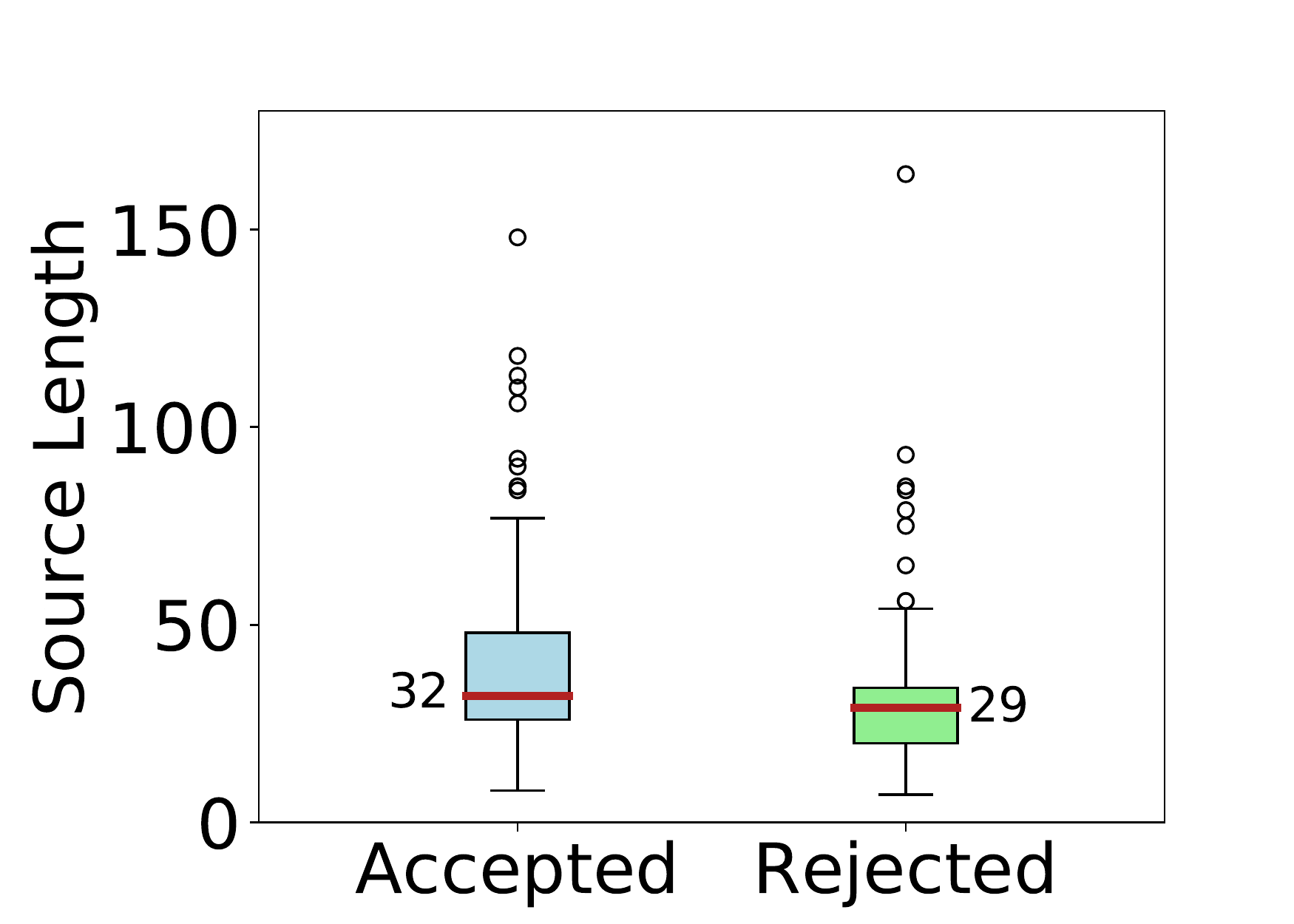}
    \caption{}
    \label{fig:src_len_acc_rej}
\end{subfigure}
\begin{subfigure}[b]{0.24\textwidth}
    \centering
    \includegraphics[width=\textwidth]{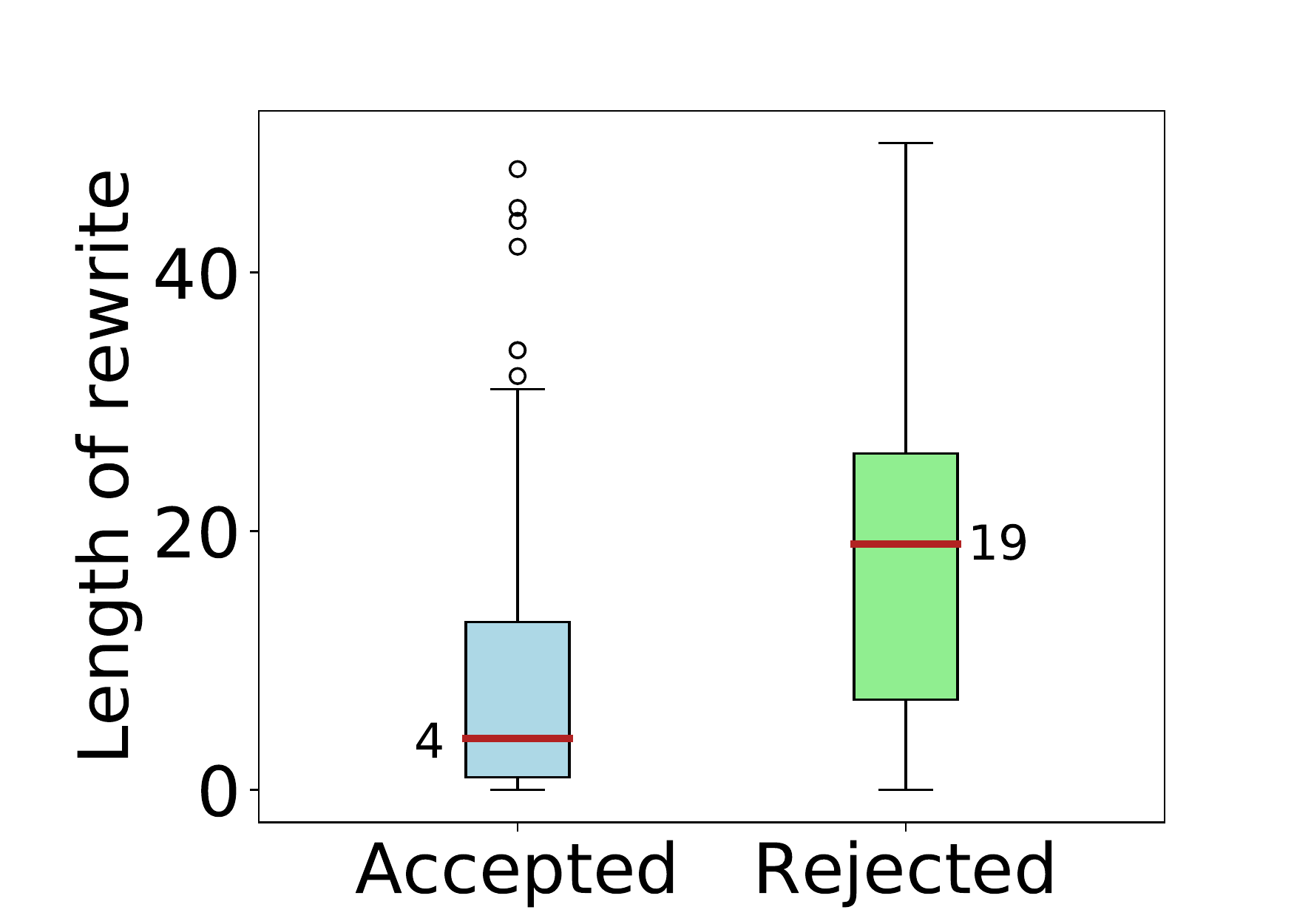}
    \caption{}
    \label{fig:infill_len_acc_rej}
\end{subfigure}\hspace{5mm}\\
\begin{subfigure}[b]{0.24\textwidth}
    \centering
    \includegraphics[width=\textwidth]{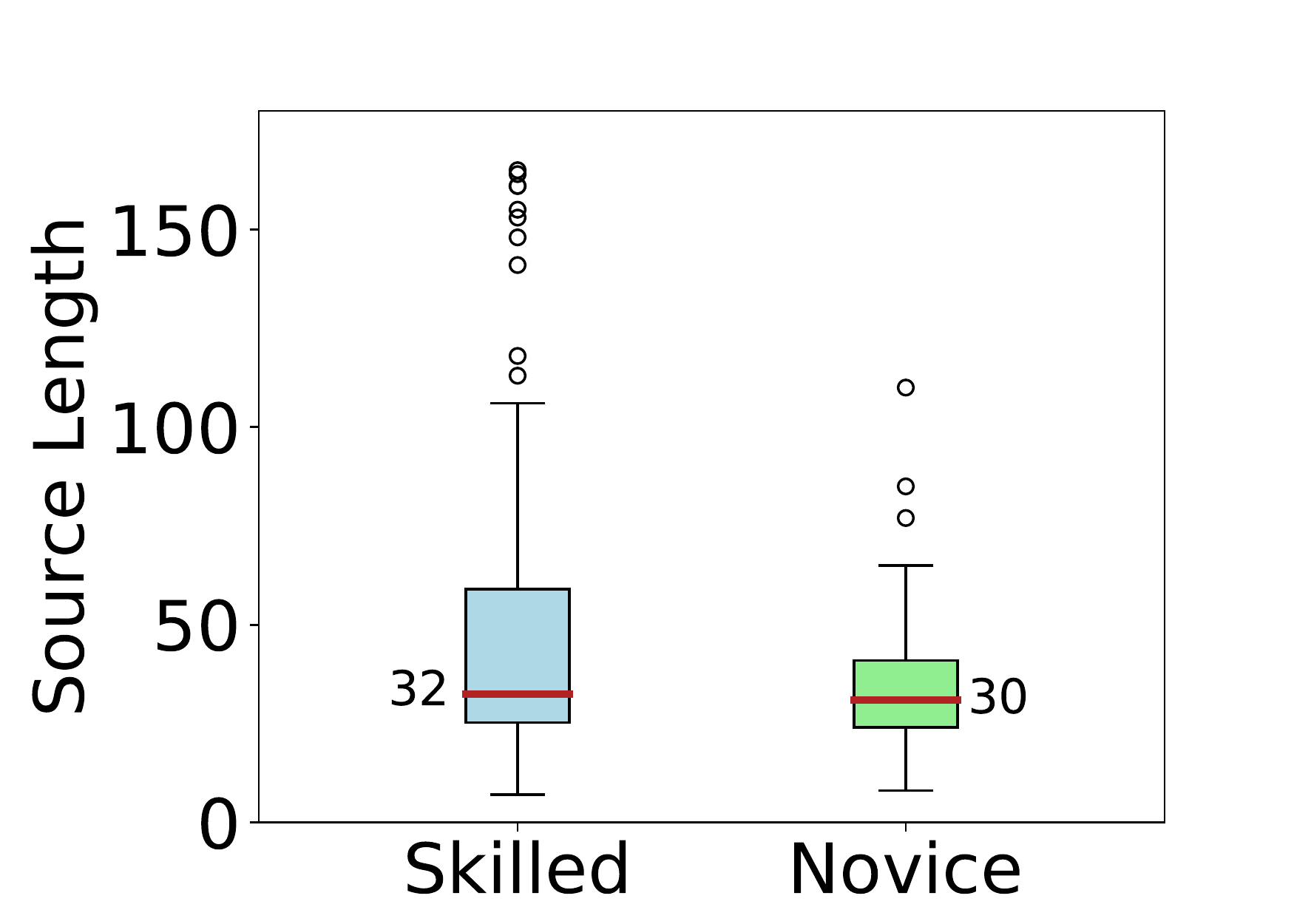}
    \caption{}
    \label{fig:src_len_exp_nov}
\end{subfigure}
\begin{subfigure}[b]{0.24\textwidth}
    \centering
    \includegraphics[width=\textwidth]{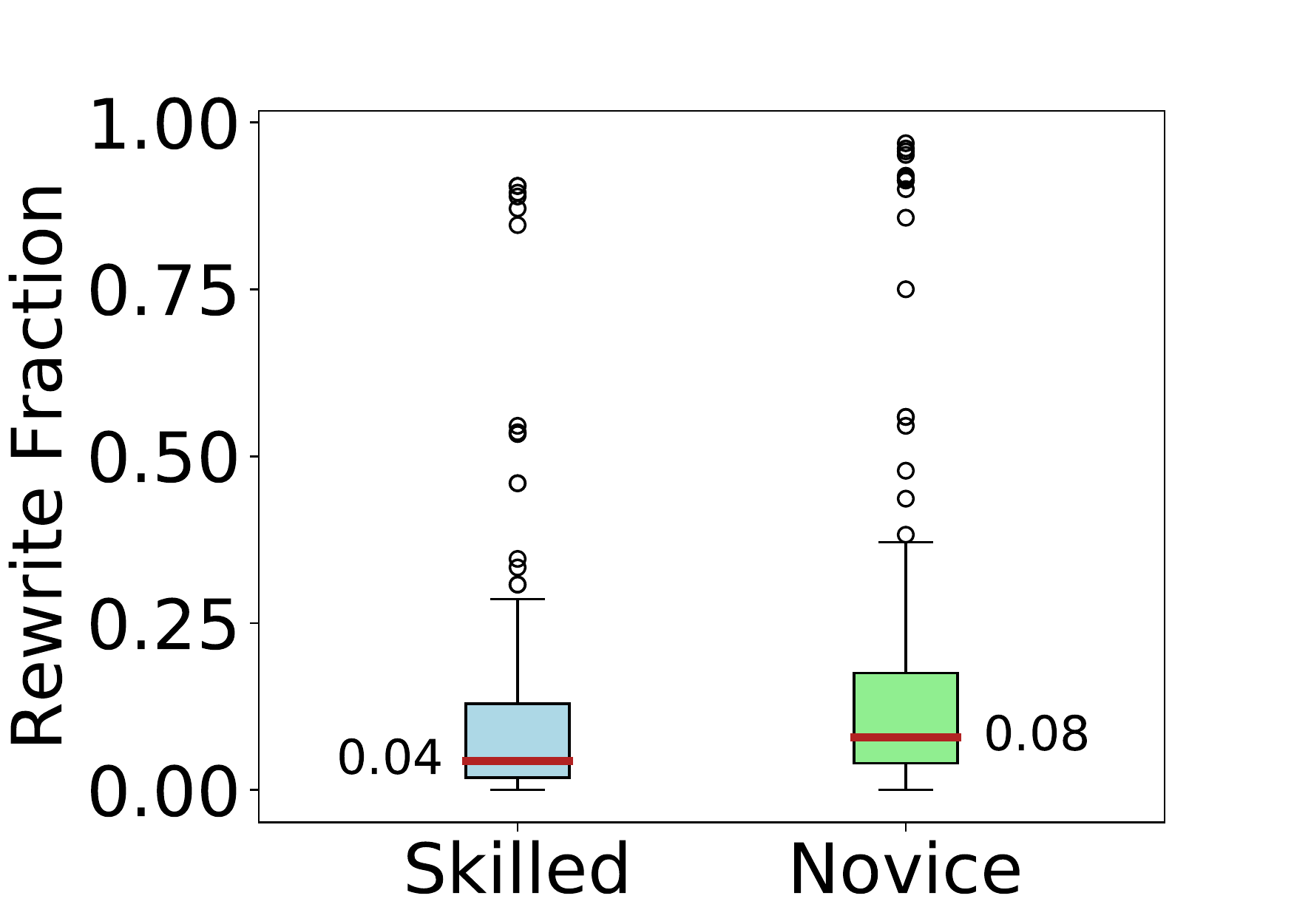}
    \caption{}
    \label{fig:frac_high_exp_nov}
\end{subfigure}
\caption{
Analysis of interactions in terms of length of source sentences provided to the model (a, c) and rewritten spans in the generated text (b, d). In each boxplot, the box indicates the interquartile range with the median values marked by the red line. Length is measured in terms of number of characters.
We see that the model is more effective when given longer source context sentences (a) and generating smaller rewritten spans of text in the target sentences (b). Skilled writers find the model to be more effective (\Cref{tab:sa_breakdown}) because they play to the model's strengths by writing longer context sentences (c) and requesting shorter spans to be rewritten in them (d).
}
\end{figure}

\paragraph{Why is \modelacc more helpful to skilled users?} %
The model is more effective at editing longer sentences.
A longer context allows the model to better infer the content and style of the requested suggestion, so we expect that the model would be more effective at editing longer sentences.
In \Cref{fig:src_len_acc_rej}, we see that the accepted suggestions are indeed generated from longer source sentences compared to the rejected ones. From \Cref{fig:src_len_exp_nov}, we also see that skilled writers tend to write longer sentences (which \modelacc is good at); this partially explains why skilled users find the model to be more helpful.
\Cref{fig:frac_high_exp_nov} also shows us that though skilled writers tend to write longer sentences, they request smaller fractions of these sentences to be rewritten. Examples 1 and 2 in \Cref{tab:pos_examples} are representative of this scenario where the model provides helpful suggestions. 

\paragraph{Takeaways} This finding of disproportionate assistance highlights a need for careful, user-centric study of interactive systems as they become more ubiquitous. Given that the use of technology could widen the gap in performance between different users, future direction to explore include developing models that assist both sets of users equally and also developing evaluation metrics that capture how performance varies across users.

\section{Related Work and Discussion}

\paragraph{Collaborative writing.}
Our work builds on existing literature on collaborative writing. Early approaches \cite{swanson2012say, roemmele2015creative} that provide text suggestions to users in the creative domain were retrieval-based.
\emph{Creative Help} \cite{roemmele2015creative} retrieved sentence-level suggestions at locations specified by a user from a large corpus of stories. A follow-up study \cite{roemmele2018linguistic} 
found that grammaticality and the presence of noun phrases in the text were indicative of helpful suggestions. 
We observe similar trends in \Cref{sec:analysis} and \Cref{app:pos}. More recently, collaborative systems have incorporated text generation models for assistance. \citet{clark2018creative} evaluated a \mil setting on the tasks of story and slogan writing. 
They tested one system that generates 
sentence-level continuations for story writing and another one that generates
a slogan from a given set of keywords, and found that solo-writing was a very competitive baseline.
\citet{akoury2020storium} gave human writers a machine-generated 
draft for storytelling and observed that writers tend to retain only a fraction of the generated text. 
\citet{coenen2021wordcraft} frames collaborative writing as a conversation between a human and a dialog system leveraging large language models. Our work is closest to \citet{ito-etal-2020-langsmith}, which demonstrated that a collaborative rewriting system helps non-native English speakers revise drafts of research papers. We focus on the more challenging domain of creative writing where users are more selective of the suggestions they accept. In addition, we study how the assistant helps with the creating writing process in an end-to-end manner, whereas \citet{ito-etal-2020-langsmith} focus on editing a given draft. 

\paragraph{Editing models.} Transformer models have shown to be good at editing text to change the style \cite{shih2019xl, krishna2020style}, debias text \cite{ma2020powertransformer}, post-edit translations \cite{grangier-auli-2018-quickedit, wang-etal-2020-touch} and simplify text \cite{kumar2020iterative}. 
\citet{chakrabarty-etal-2021-mermaid} train a model to generate metaphors employing a pseudo-parallel corpus of metaphoric sentences and corresponding literal sentences similar to how we use the sources of creative text.
Additionally, infilling literature \cite{donahue2020infilling, fedus2018maskgan, joshi2019spanbert, shen2020blank} has shown that we can train models to fill in blanks. 
We incorporate editing models to collaborative writing which adapts to human feedback.

\section{Conclusions and Future Work}
In this paper, we develop 
a Creative Rewriting Assistant that is able to effectively assist users to 
complete the task of creative image captioning. Our \mil rewriting setup allows  
human users to control the content of their writing while  
utilizing the strengths of text generation models.  
Our model is found to be more useful for skilled users,
so it remains to be explored how to better assist novice writers. 
One direction is to explore generating text from keywords because these users might need help with planning the global content and structure of their writing. 
Additionally, 
the most common error mode we see amongst the rejected suggestions is content drift, 
so another challenge 
is to balance faithfulness to the author's content 
with 
creativity in text generation. 
\section*{Acknowledgements}
We would like to thank the anonymous reviewers for their helpful comments. This work is supported by the National Science Foundation under Grant No. 1922658 and the Samsung Advanced Institute of Technology (Next Generation Deep Learning: From Pattern Recognition to AI). 

\clearpage
\newpage
\bibliography{anthology,custom,all}

\begin{thebibliography}{32}
\expandafter\ifx\csname natexlab\endcsname\relax\def\natexlab#1{#1}\fi

\bibitem[{Akoury et~al.(2020)Akoury, Wang, Whiting, Hood, Peng, and
  Iyyer}]{akoury2020storium}
Nader Akoury, Shufan Wang, Josh Whiting, Stephen Hood, Nanyun Peng, and Mohit
  Iyyer. 2020.
\newblock \href {https://doi.org/10.18653/v1/2020.emnlp-main.525} {{STORIUM}:
  {A} {D}ataset and {E}valuation {P}latform for {M}achine-in-the-{L}oop {S}tory
  {G}eneration}.
\newblock In \emph{Proceedings of the 2020 Conference on Empirical Methods in
  Natural Language Processing (EMNLP)}, pages 6470--6484, Online. Association
  for Computational Linguistics.

\bibitem[{Bostan et~al.(2020)Bostan, Kim, and
  Klinger}]{bostan2020goodnewseveryone}
Laura Ana~Maria Bostan, Evgeny Kim, and Roman Klinger. 2020.
\newblock \href {https://www.aclweb.org/anthology/2020.lrec-1.194}
  {{G}ood{N}ews{E}veryone: A corpus of news headlines annotated with emotions,
  semantic roles, and reader perception}.
\newblock In \emph{Proceedings of the 12th Language Resources and Evaluation
  Conference}, pages 1554--1566, Marseille, France. European Language Resources
  Association.

\bibitem[{Chakrabarty et~al.(2021)Chakrabarty, Zhang, Muresan, and
  Peng}]{chakrabarty-etal-2021-mermaid}
Tuhin Chakrabarty, Xurui Zhang, Smaranda Muresan, and Nanyun Peng. 2021.
\newblock \href {https://doi.org/10.18653/v1/2021.naacl-main.336} {{MERMAID}:
  Metaphor generation with symbolism and discriminative decoding}.
\newblock In \emph{Proceedings of the 2021 Conference of the North American
  Chapter of the Association for Computational Linguistics: Human Language
  Technologies}, pages 4250--4261, Online. Association for Computational
  Linguistics.

\bibitem[{Chen et~al.(2015)Chen, Kuznetsova, Warren, and Choi}]{chen2015deja}
Jianfu Chen, Polina Kuznetsova, David Warren, and Yejin Choi. 2015.
\newblock \href {https://doi.org/10.3115/v1/N15-1053} {D{\'e}j{\`a}
  image-captions: A corpus of expressive descriptions in repetition}.
\newblock In \emph{Proceedings of the 2015 Conference of the North {A}merican
  Chapter of the Association for Computational Linguistics: Human Language
  Technologies}, pages 504--514, Denver, Colorado. Association for
  Computational Linguistics.

\bibitem[{Clark et~al.(2018)Clark, Ross, Tan, Ji, and
  Smith}]{clark2018creative}
Elizabeth Clark, Anne~Spencer Ross, Chenhao Tan, Yangfeng Ji, and Noah~A Smith.
  2018.
\newblock Creative writing with a machine in the loop: Case studies on slogans
  and stories.
\newblock In \emph{23rd International Conference on Intelligent User
  Interfaces}, pages 329--340.

\bibitem[{Coenen et~al.(2021)Coenen, Davis, Ippolito, Reif, and
  Yuan}]{coenen2021wordcraft}
Andy Coenen, Luke Davis, Daphne Ippolito, Emily Reif, and Ann Yuan. 2021.
\newblock \href {http://arxiv.org/abs/2107.07430} {Wordcraft: a human-ai
  collaborative editor for story writing}.
\newblock \emph{CoRR}, abs/2107.07430.

\bibitem[{Donahue et~al.(2020)Donahue, Lee, and Liang}]{donahue2020infilling}
Chris Donahue, Mina Lee, and Percy Liang. 2020.
\newblock Enabling language models to fill in the blanks.
\newblock In \emph{Association for Computational Linguistics (ACL)}.

\bibitem[{Fedus et~al.(2018)Fedus, Goodfellow, and Dai}]{fedus2018maskgan}
William Fedus, Ian Goodfellow, and Andrew~M. Dai. 2018.
\newblock Maskgan: Better text generation via filling in the.
\newblock In \emph{International Conference on Learning Representations
  (ICLR)}.

\bibitem[{Garfield(2008)}]{garfield2008creativity}
Monica~J Garfield. 2008.
\newblock Creativity support systems.
\newblock In \emph{Handbook on Decision Support Systems 2}, pages 745--758.
  Springer.

\bibitem[{Gordon et~al.(2015)Gordon, Hobbs, May, Mohler, Morbini, Rink,
  Tomlinson, and Wertheim}]{gordon2015corpus}
Jonathan Gordon, Jerry Hobbs, Jonathan May, Michael Mohler, Fabrizio Morbini,
  Bryan Rink, Marc Tomlinson, and Suzanne Wertheim. 2015.
\newblock \href {https://doi.org/10.3115/v1/W15-1407} {A corpus of rich
  metaphor annotation}.
\newblock In \emph{Proceedings of the Third Workshop on Metaphor in {NLP}},
  pages 56--66, Denver, Colorado. Association for Computational Linguistics.

\bibitem[{Grangier and Auli(2018)}]{grangier-auli-2018-quickedit}
David Grangier and Michael Auli. 2018.
\newblock \href {https://doi.org/10.18653/v1/N18-1025} {{Q}uick{E}dit: Editing
  text {\&} translations by crossing words out}.
\newblock In \emph{Proceedings of the 2018 Conference of the North {A}merican
  Chapter of the Association for Computational Linguistics: Human Language
  Technologies, Volume 1 (Long Papers)}, pages 272--282, New Orleans,
  Louisiana. Association for Computational Linguistics.

\bibitem[{Holtzman et~al.(2019)Holtzman, Buys, Du, Forbes, and
  Choi}]{holtzman2019curious}
Ari Holtzman, Jan Buys, Li~Du, Maxwell Forbes, and Yejin Choi. 2019.
\newblock The curious case of neural text degeneration.
\newblock In \emph{International Conference on Learning Representations}.

\bibitem[{Ito et~al.(2020)Ito, Kuribayashi, Hidaka, Suzuki, and
  Inui}]{ito-etal-2020-langsmith}
Takumi Ito, Tatsuki Kuribayashi, Masatoshi Hidaka, Jun Suzuki, and Kentaro
  Inui. 2020.
\newblock \href {https://doi.org/10.18653/v1/2020.emnlp-demos.28} {Langsmith:
  An interactive academic text revision system}.
\newblock In \emph{Proceedings of the 2020 Conference on Empirical Methods in
  Natural Language Processing: System Demonstrations}, pages 216--226, Online.
  Association for Computational Linguistics.

\bibitem[{Joshi et~al.(2019)Joshi, Chen, Liu, Weld, Zettlemoyer, and
  Levy}]{joshi2019spanbert}
Mandar Joshi, Danqi Chen, Yinhan Liu, Daniel~S. Weld, Luke Zettlemoyer, and
  Omer Levy. 2019.
\newblock Span{BERT}: Improving pre-training by representing and predicting
  spans.
\newblock \emph{arXiv preprint arXiv:1907.10529}.

\bibitem[{Kingma and Ba(2014)}]{kingma2014adam}
Diederik~P Kingma and Jimmy Ba. 2014.
\newblock Adam: A method for stochastic optimization.
\newblock \emph{arXiv preprint arXiv:1412.6980}.

\bibitem[{Krishna et~al.(2020)Krishna, Wieting, and Iyyer}]{krishna2020style}
Kalpesh Krishna, Josh Wieting, and Mohit Iyyer. 2020.
\newblock Reformulating unsupervised style transfer as paraphrase generation.
\newblock In \emph{Empirical Methods in Natural Language Processing}.

\bibitem[{Kumar et~al.(2020)Kumar, Mou, Golab, and
  Vechtomova}]{kumar2020iterative}
Dhruv Kumar, Lili Mou, Lukasz Golab, and Olga Vechtomova. 2020.
\newblock \href {https://doi.org/10.18653/v1/2020.acl-main.707} {Iterative
  edit-based unsupervised sentence simplification}.
\newblock In \emph{Proceedings of the 58th Annual Meeting of the Association
  for Computational Linguistics}, pages 7918--7928, Online. Association for
  Computational Linguistics.

\bibitem[{Lewis et~al.(2019)Lewis, Liu, Goyal, Ghazvininejad, Mohamed, Levy,
  Stoyanov, and Zettlemoyer}]{lewis2019bart}
Mike Lewis, Yinhan Liu, Naman Goyal, Marjan Ghazvininejad, Abdelrahman Mohamed,
  Omer Levy, Ves Stoyanov, and Luke Zettlemoyer. 2019.
\newblock Bart: Denoising sequence-to-sequence pre-training for natural
  language generation, translation, and comprehension.
\newblock \emph{arXiv preprint arXiv:1910.13461}.

\bibitem[{Ma et~al.(2020)Ma, Sap, Rashkin, and Choi}]{ma2020powertransformer}
Xinyao Ma, Maarten Sap, Hannah Rashkin, and Yejin Choi. 2020.
\newblock Powertransformer: Unsupervised controllable revision for biased
  language correction.
\newblock In \emph{EMNLP}.

\bibitem[{Mohammad et~al.(2016)Mohammad, Shutova, and
  Turney}]{mohammad2016metaphor}
Saif Mohammad, Ekaterina Shutova, and Peter Turney. 2016.
\newblock \href {https://doi.org/10.18653/v1/S16-2003} {Metaphor as a medium
  for emotion: An empirical study}.
\newblock In \emph{Proceedings of the Fifth Joint Conference on Lexical and
  Computational Semantics}, pages 23--33, Berlin, Germany. Association for
  Computational Linguistics.

\bibitem[{Mohler et~al.(2015)Mohler, Tomlinson, and Rink}]{mohler2015cross}
Michael Mohler, Marc~T Tomlinson, and Bryan Rink. 2015.
\newblock Cross-lingual semantic generalization for the detection of metaphor.
\newblock \emph{Computational Linguistics and Intelligent Text Processing}.

\bibitem[{Niculae and Danescu-Niculescu-Mizil(2014)}]{niculae2014brighter}
Vlad Niculae and Cristian Danescu-Niculescu-Mizil. 2014.
\newblock \href {https://doi.org/10.3115/v1/D14-1215} {Brighter than gold:
  Figurative language in user generated comparisons}.
\newblock In \emph{Proceedings of the 2014 Conference on Empirical Methods in
  Natural Language Processing ({EMNLP})}, pages 2008--2018, Doha, Qatar.
  Association for Computational Linguistics.

\bibitem[{Ott et~al.(2019)Ott, Edunov, Baevski, Fan, Gross, Ng, Grangier, and
  Auli}]{ott2019fairseq}
Myle Ott, Sergey Edunov, Alexei Baevski, Angela Fan, Sam Gross, Nathan Ng,
  David Grangier, and Michael Auli. 2019.
\newblock fairseq: A fast, extensible toolkit for sequence modeling.
\newblock In \emph{Proceedings of NAACL-HLT 2019: Demonstrations}.

\bibitem[{Roemmele and Gordon(2018)}]{roemmele2018linguistic}
Melissa Roemmele and Andrew Gordon. 2018.
\newblock \href {https://doi.org/10.18653/v1/W18-1502} {Linguistic features of
  helpfulness in automated support for creative writing}.
\newblock In \emph{Proceedings of the First Workshop on Storytelling}, pages
  14--19, New Orleans, Louisiana. Association for Computational Linguistics.

\bibitem[{Roemmele and Gordon(2015)}]{roemmele2015creative}
Melissa Roemmele and Andrew~S Gordon. 2015.
\newblock Creative help: A story writing assistant.
\newblock In \emph{International Conference on Interactive Digital
  Storytelling}, pages 81--92. Springer.

\bibitem[{Samuel et~al.(2016)Samuel, Mateas, and
  Wardrip-Fruin}]{samuel2016design}
Ben Samuel, Michael Mateas, and Noah Wardrip-Fruin. 2016.
\newblock The design of writing buddy: a mixed-initiative approach towards
  computational story collaboration.
\newblock In \emph{International Conference on Interactive Digital
  Storytelling}, pages 388--396. Springer.

\bibitem[{Shen et~al.(2020)Shen, Quach, Barzilay, and Jaakkola}]{shen2020blank}
Tianxiao Shen, Victor Quach, Regina Barzilay, and Tommi Jaakkola. 2020.
\newblock Blank language models.
\newblock In \emph{Proceedings of the 2020 Conference on Empirical Methods in
  Natural Language Processing (EMNLP)}, pages 5186--5198.

\bibitem[{Shih et~al.(2019)Shih, Chang, and Yang}]{shih2019xl}
Yong-Siang Shih, Wei-Cheng Chang, and Yiming Yang. 2019.
\newblock {XL}-{E}ditor: Post-editing sentences with xlnet.
\newblock \emph{arXiv preprint arXiv:1910.10479}.

\bibitem[{Steen et~al.(2010)Steen, Dorst, Herrmann, Kaal, Krennmayr, and
  Pasma}]{vuamc}
G.J. Steen, A.G. Dorst, J.B. Herrmann, A.A. Kaal, T.~Krennmayr, and T.~Pasma.
  2010.
\newblock \emph{A method for linguistic metaphor identification. From MIP to
  MIPVU.}
\newblock Number~14 in Converging Evidence in Language and Communication
  Research. John Benjamins.

\bibitem[{Swanson and Gordon(2012)}]{swanson2012say}
Reid Swanson and Andrew~S. Gordon. 2012.
\newblock \href {https://doi.org/10.1145/2362394.2362398} {Say anything: Using
  textual case-based reasoning to enable open-domain interactive storytelling}.
\newblock \emph{ACM Trans. Interact. Intell. Syst.}, 2(3).

\bibitem[{Wang and Sennrich(2020)}]{wang2020exposure}
Chaojun Wang and Rico Sennrich. 2020.
\newblock \href {https://doi.org/10.18653/v1/2020.acl-main.326} {On exposure
  bias, hallucination and domain shift in neural machine translation}.
\newblock In \emph{Proceedings of the 58th Annual Meeting of the Association
  for Computational Linguistics}, pages 3544--3552, Online. Association for
  Computational Linguistics.

\bibitem[{Wang et~al.(2020)Wang, Zhang, Liu, Huang, and
  Zong}]{wang-etal-2020-touch}
Qian Wang, Jiajun Zhang, Lemao Liu, Guoping Huang, and Chengqing Zong. 2020.
\newblock \href {https://aclanthology.org/2020.aacl-main.1} {Touch editing: A
  flexible one-time interaction approach for translation}.
\newblock In \emph{Proceedings of the 1st Conference of the Asia-Pacific
  Chapter of the Association for Computational Linguistics and the 10th
  International Joint Conference on Natural Language Processing}, pages 1--11,
  Suzhou, China. Association for Computational Linguistics.

\end{thebibliography}
\bibliographystyle{acl_natbib}
\clearpage
\newpage
\appendix
\section{Ethical Considerations}

\paragraph{Disproportionate assistance.} One of the findings of our work was that the collaboration model discussed is more effective at assisting users who are already skilled at writing tasks. We noted in the paper that an important direction of future work is to develop systems that cater to the novice user group as well. An ethical consideration is that if such a system in its current state were deployed, it could lead to an increase in the disparity in performance between the two user groups. We believe that recording this observation is important as human-centered machine learning systems become more prevalent.

\paragraph{Appropriate remuneration for crowd workers.} To complete the HIT on AMT, workers need to interact with the model a minimum of 2 times before submitting the caption---it is explicitly mentioned that they are free to reject the suggestions and accepting/rejecting suggestions has no bearing on the payment. From a small internal pilot (also confirmed with Mechanical Turk experiments) we estimate an average completion time to be 10 minutes with an additional 2 minutes to read the instructions, so the payment is set to \$3 for the HIT (prorated to an hourly wage of \$15). The estimated completion time for third-party evaluation was 1 minute so the payment was set to \$0.25 per annotation (prorated to an hourly wage of \$15).

\section{HIT Instructions and Details}
\label{sec:hit_instructions}

\begin{figure*}[ht!]
    \centering
    \includegraphics[width=0.7\textwidth]{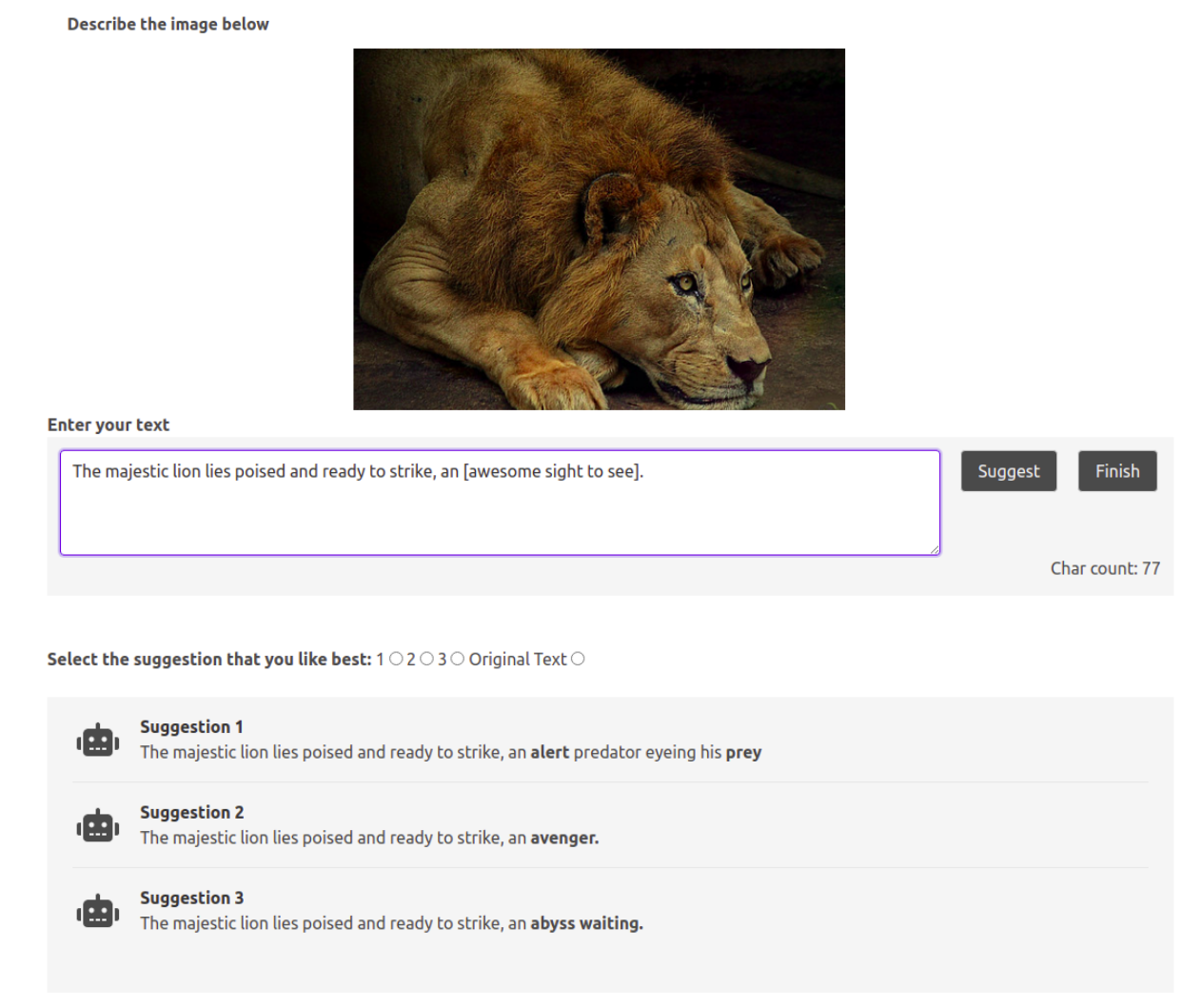}
    \caption{User interface. The user demarcates the span they want suggestions for in a text box and the model offers three suggestions for the user to pick from. This continues iteratively till the human is satisfied and submits the caption to finish the task.}
    \label{fig:interface}
\end{figure*}
\Cref{fig:interface} is a screenshot of the interface presented to the crowdworkers for the writing task.

\subsection{Instructions for crowdworkers completing the writing task}
\begin{itemize}
    \item Along with the first question in the survey is a link to the image captioning task. Navigate there. You will see a panel on the top left that shows you an image that you need to describe.
    \item You're free to interpret the image as you please---be as descriptive/figurative as possible. 
    \item To help you with the same, we have a feature where you can highlight a part of your text with square brackets (`[', `]') and request targeted suggestions in that area. Please look at the accompanying examples on how to use it effectively.
    \item While writing we find that we are often able to provide content but to make the text more interesting is difficult, hopefully the assistant helps there. You will always have the option to reject the suggestions of the assistant and switch back to your original text. Bear in mind that the assistant isn't really great at guessing content words.
    \item To complete the task, continue editing until you are happy with the description. We require that you at least request suggestions from the assistant for a minimum of two times, even if you choose to reject the suggestions.
\end{itemize}

\subsection{Instructions for crowdworkers evaluating the captions}
\begin{itemize}
    \item Choose the better (more descriptive and/or figurative) caption for the image.
    \item A better caption is your subjective judgement, the rubrics to make the choice are that the caption is descriptive and/or figurative in its interpretation of the image (Refer the examples for further clarification).
    \item The explanation asked is supposed to be very brief. A single word of if you like it for being descriptive or interpretive will do.
    \item Relevance of the caption to the image is your subjective choice whether the caption appropriately represents what is in the image and is not just a catchy piece of text unrelated to the image.
    \item A caption that you deem irrelevant should never be the better caption, unless both are irrelevant.
\end{itemize}

\section{User Feedback from Mechanical Turk}
\label{sec:user_feedback}
We present some user feedback obtained from the task---these cover some of the positive and negative comments we received. The negative comments are representative of some of the issues we highlight in \cref{sec:error_modes}
\paragraph{Positive}
\begin{itemize}
    \item I was impressed by how well this worked. I feel like my writing did improve by using the suggestions. At the very least it gave me good ideas.
    \item I got great suggestions that offered me words that I hadn't considered and fit even better than my own writing so I was pleased with the suggestions.
    \item I think everything was clear and straightforward and I enjoyed the interface.
\end{itemize}
\paragraph{Negative}
\begin{itemize}
    \item The suggestions were sometimes too far from the meaning of the original text so that they no longer made sense or were not grammatically correct.
    \item The instructions were fine, but the suggestions sure leave a lot to be desired.  It replaced 'bright yellow' with red a couple of times.
\end{itemize}

\section{Details for Reproducibility Checklist}

\subsection{Model Details} 
We use a pre-trained BART-Large (406M parameters) model as the starting point for our experiments, which was made available through the \texttt{fairseq} \cite{ott2019fairseq} implementation. Unless mentioned otherwise, the recommended values for the hyperparameters were taken from the released 
fine tuning script in the library. We selected the learning rate for \Cref{sec:user_eval} using validation perplexity as a metric varying the value from $1 \times 10^{-5}$ to $1 \times 10^{-4}$. The source code for our experiments, both to set up the interface and train the model, will be made available upon publication of this work.
Model training was on a Titan Xp single GPU machine with 12GB of memory. The same machine was also used to host the server for the interactive experiments. A model inference is made for each request from the users. 

\subsection{Data Details}
All the datasets from \Cref{tab:sources} are publicly available already. As highlighted in \Cref{sec:analysis}, one reason our model suffers from content drift is because the creation process does not guarantee that the content in the source and target is identical. So prior to making the pseudo-parallel corpus from \Cref{sec:approach} available, we aim to filter out those examples which suffer from content drift. The dataset of interactions from \Cref{sec:learning_from_interactions} cannot be directly shared.

\section{Further Analysis}

\paragraph{Longer model rewrites get rejected more frequently.}

Our assumption is that users want to control the content of the caption. 
When the model rewrites a longer span and adds more new text to the draft, it is likely to diverge from the original content given by the user. 
We compare the length of new  
text introduced into the draft by \modelacc in both the accepted and rejected suggestions. From \Cref{fig:infill_len_acc_rej}, we  see that  longer revisions are more likely to be rejected. 

\subsection{Collaborative vs Human Writing}
\begin{figure}[ht!]
\centering
\begin{subfigure}[b]{0.24\textwidth}
    \centering
    \includegraphics[width=\textwidth]{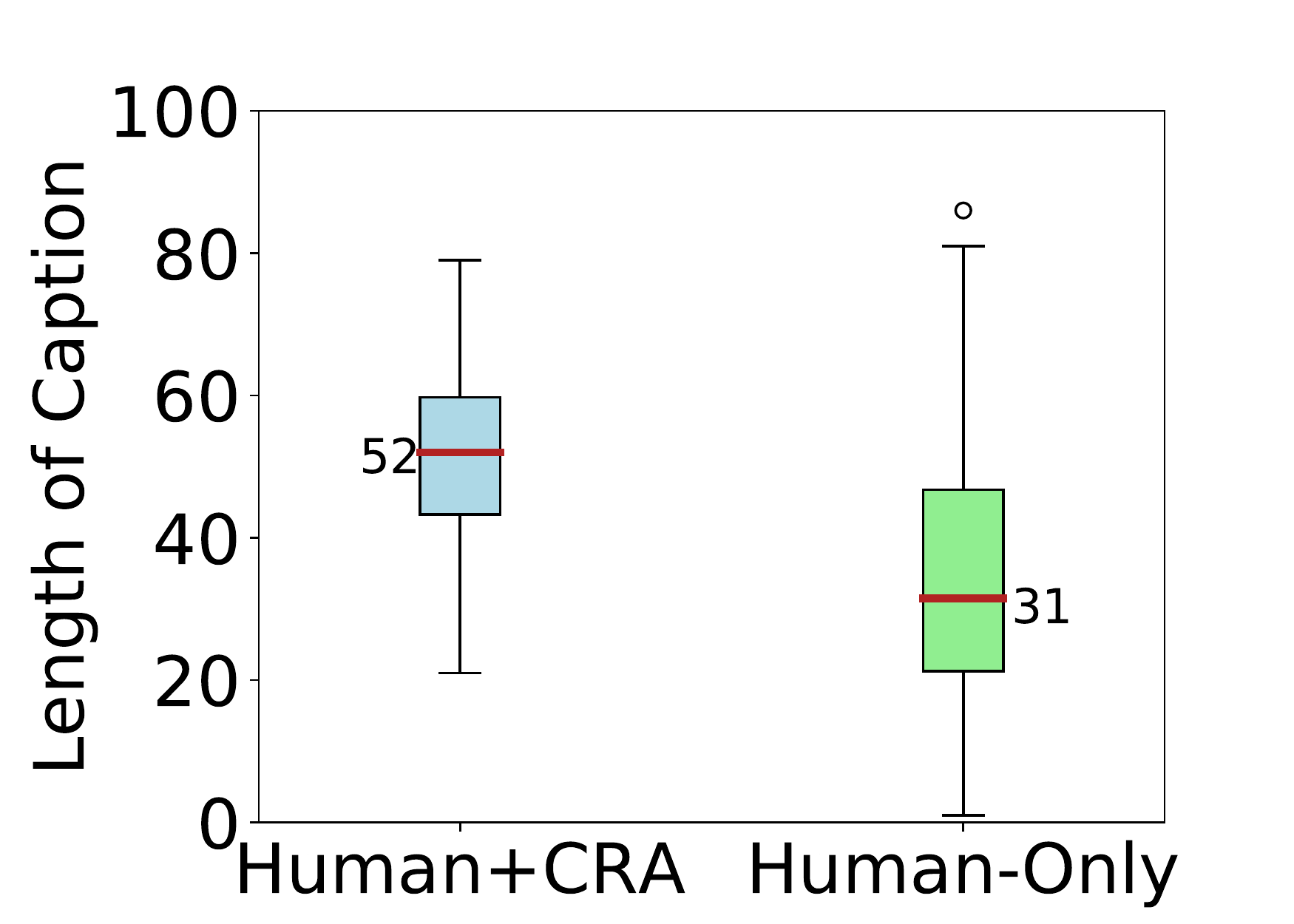}
    \caption{}
    \label{fig:app_cra_human_len}
\end{subfigure}
\begin{subfigure}[b]{0.24\textwidth}
    \centering
    \includegraphics[width=\textwidth]{Figures/cra_human_ngrams.pdf}
    \caption{}
    \label{fig:app_cra_human_ngram}
\end{subfigure}\hspace{5mm}\\
\begin{subfigure}[b]{0.24\textwidth}
    \centering
    \includegraphics[width=\textwidth]{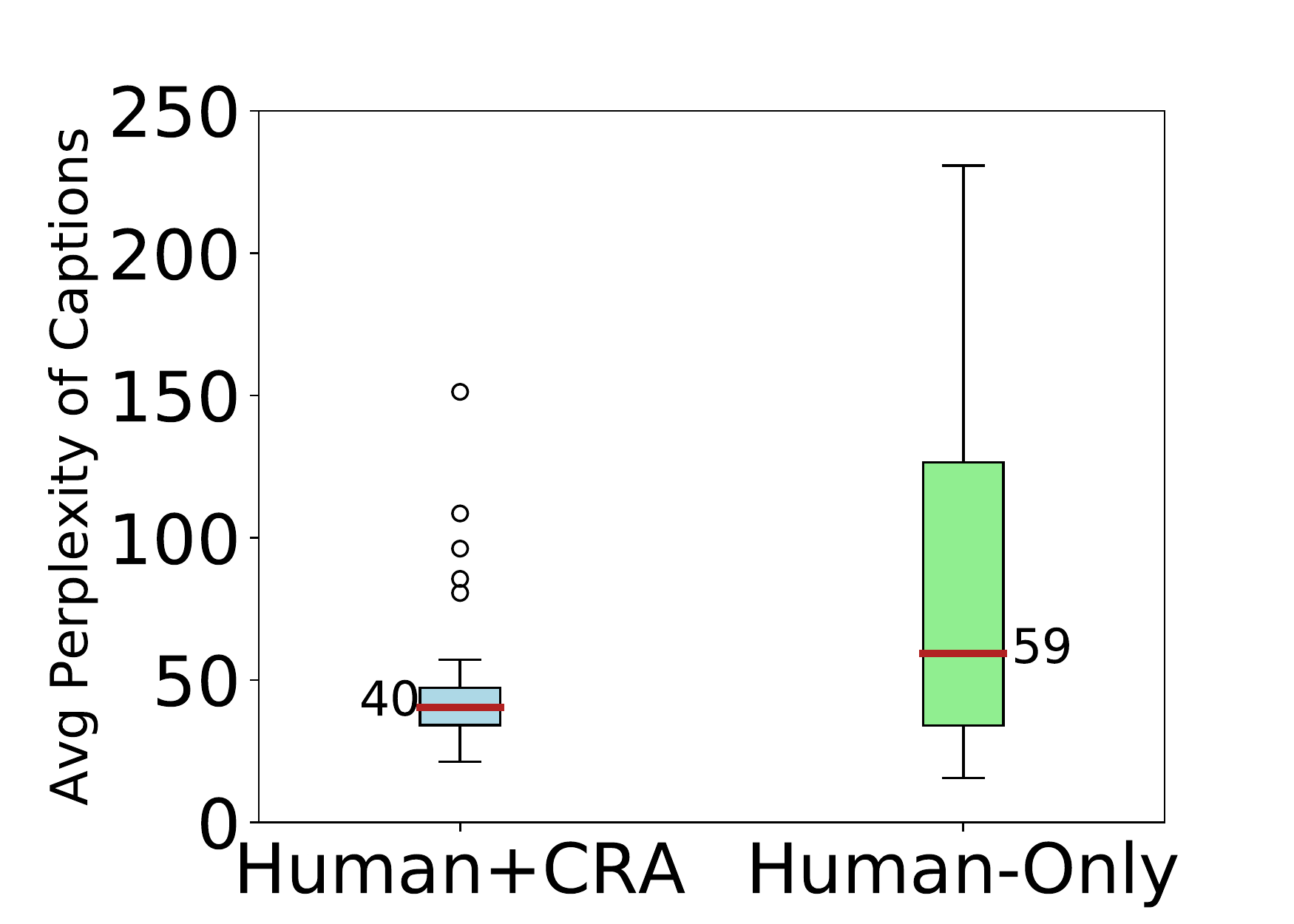}
    \caption{}
    \label{fig:app_cra_human_ppl}
\end{subfigure}
\caption{Comparison of text generated from a collaborative setup (Human+\modelacc) and solo-writers (Human-Only). We see that collaborative users tend to write longer captions (\Cref{fig:app_cra_human_len}), that contain more unique N-grams (\Cref{fig:app_cra_human_ngram}, N=3), and on average have a lower perplexity (\Cref{fig:app_cra_human_ppl}), as evaluated using a pre-trained GPT2 model. We use perplexity as a proxy for fluency in text. Collaborative users tend to consider more diverse options for text while retaining fluency in the text.}
\end{figure}

In \Cref{sec:third_party_eval}, we saw that humans writing in a collaborative setup tend to produce better creative output. To analyze how model intervention affects the text, we collected some statistics on the 100 final captions produced from the Human+\modelacc setup and the Human-Only setup. We see that users in the collaborative setup write longer captions (\Cref{fig:app_cra_human_len}) that tend to have more unique n-grams (\Cref{fig:cra_human_ngram}), indicative that users are incorporating more diverse elements into their text as a result of model interaction. 
We also calculate the perplexity of the final captions using a pre-trained GPT2 model. 
From \Cref{fig:app_cra_human_ppl}, we see that the captions with \modelacc intervention have a lower average perplexity despite having higher lexical diversity. Also we see that the perplexity scores from the collaborative setup have significantly less variance than the human-only captions indicating that collaboration makes different people's writing more similar to each other.

\subsection{POS Tags}
\label{app:pos}
\begin{figure}[ht!]
    \begin{subfigure}[b]{0.5\textwidth}
        \centering
        \includegraphics[width=8cm]{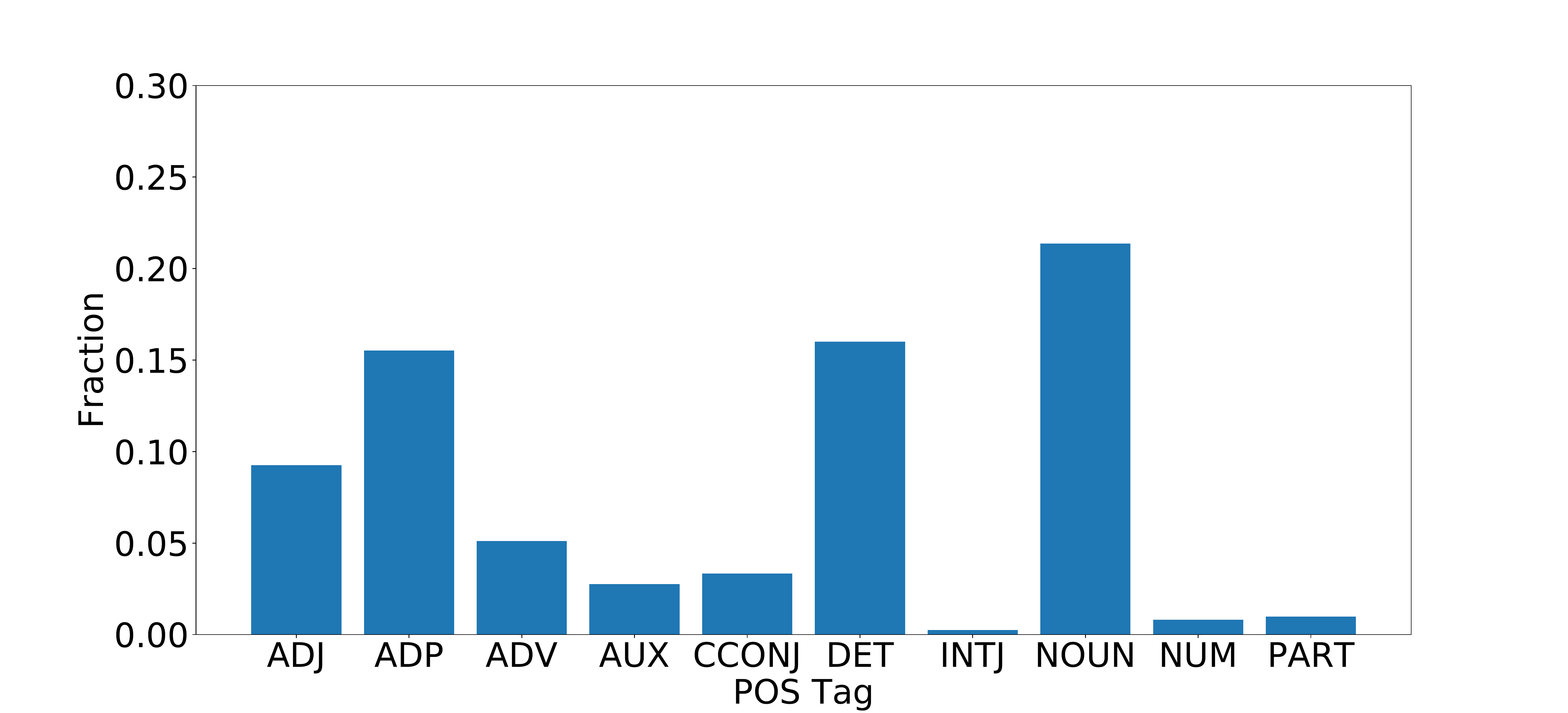}
        \caption{POS tags of rewritten text for all accepted suggestions.}
        \label{fig:infill_pos_acc}    
    \end{subfigure}
    \\
    \begin{subfigure}[b]{0.5\textwidth}
        \centering
        \includegraphics[width=8cm]{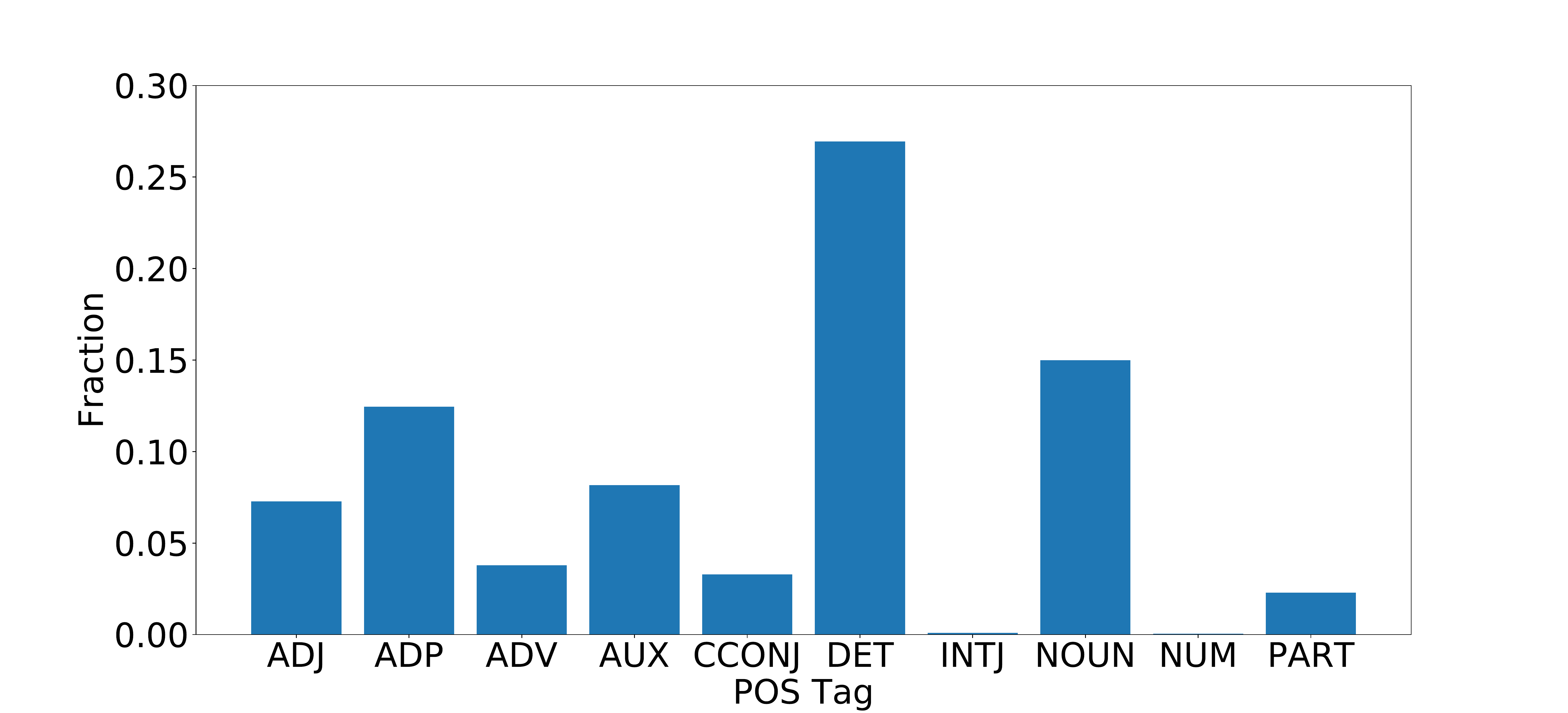}
        \caption{POS tags of rewritten text for all rejected suggestions.}
        \label{fig:infill_pos_rej}    
    \end{subfigure}
    \caption{The 10 most common POS tags in accepted and rejected suggestions: Accepted suggestions tend to have more adjectives, adverbs and nouns and rejected suggestions tend to have higher fraction of determiners
    }
    \label{fig:infill_pos}
\end{figure}

To examine the kind of text that is helpful to users, we analyze the linguistic characteristics of accepted suggestions and rejected suggestions.
\paragraph{Accepted suggestions have more adjectives, adverbs and nouns.}
\Cref{fig:infill_pos} shows the fraction of different  POS tags in the revised span of accepted suggestions and rejected suggestions. 
Accepted suggestions tend to have a larger fraction of adverbs, adjectives and nouns, whereas rejected suggestions have a large fraction of determiners. Prior work  \cite{roemmele2018linguistic} also observed that the presence of noun phrases in suggestions has a positive correlation with helpfulness.

\end{document}